\documentclass[sigconf]{acmart}
\usepackage{pgfplots}
\pgfplotsset{compat=1.18}
\usepackage{siunitx}
\usepackage{subcaption}
\usepackage{wrapfig}
\usepackage{lipsum} % For dummy text

\copyrightyear{2025}
\acmYear{2025}
\setcopyright{acmlicensed}
\acmConference[MM '25] {Proceedings of the 33rd ACM International Conference on Multimedia}{October 27--31, 2025}{Dublin, Ireland.}
\acmBooktitle{Proceedings of the 33rd ACM International Conference on Multimedia (MM '25), October 27--31, 2025, Dublin, Ireland}
\acmDOI{10.1145/3746027.3758198}
\acmISBN{979-8-4007-2035-2/2025/10}

\begin{document}
\def\ourdataset{MSC }
\definecolor{yellow}{RGB}{255, 178, 0}
\definecolor{purple}{RGB}{114, 92, 173}
\definecolor{blue}{RGB}{140, 205, 235}

\title{\includegraphics[width=0.03\textwidth]{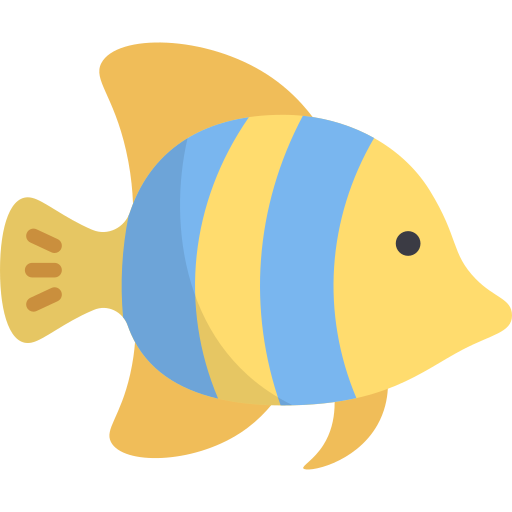}\textcolor{purple}{M}\textcolor{blue}{S}\textcolor{yellow}{C}: A \textcolor{purple}{M}arine Wildlife Video Dataset with Grounded \textcolor{blue}{S}egmentation and Clip-Level \textcolor{yellow}{C}aptioning}


\author{Quang-Trung Truong}
\affiliation{%
  \institution{The Hong Kong University of Science and Technology}
  % \streetaddress{1 Th{\o}rv{\"a}ld Circle}
  \city{Hong Kong SAR}
  \country{China}}
\email{qttruong@connect.ust.hk}

\author{Yuk-Kwan Wong}
\affiliation{%
  \institution{The Hong Kong University of Science and Technology}
  % \streetaddress{1 Th{\o}rv{\"a}ld Circle}
  \city{Hong Kong SAR}
  \country{China}}
\email{ykwongaq@connect.ust.hk}

\author{Vo Hoang Kim Tuyen Dang}
\affiliation{%
  \institution{Ho Chi Minh University of Science}
  % \streetaddress{1 Th{\o}rv{\"a}ld Circle}
  \city{Ho Chi Minh}
  \country{Viet Nam}}
\email{20120399@student.hcmus.edu.vn}

\author{Rinaldi Gotama}
\affiliation{%
  \institution{Indo Ocean Foundation}
  % \streetaddress{1 Th{\o}rv{\"a}ld Circle}
  \city{Bali}
  \country{Indonesia}}
\email{rinaldigotama@gmail.com}

\author{Duc Thanh Nguyen}
\affiliation{%
  \institution{Deakin University}
  % \streetaddress{1 Th{\o}rv{\"a}ld Circle}
  % \city{Victoria}
  \country{Australia}}
\email{duc.nguyen@deakin.edu.au}

\author{Sai-Kit Yeung}
\affiliation{%
  \institution{The Hong Kong University of Science and Technology}
  % \streetaddress{1 Th{\o}rv{\"a}ld Circle}
  \city{Hong Kong SAR}
  \country{China}}
\email{saikit@ust.hk}
\renewcommand{\shortauthors}{Quang-Trung Truong et al.}
%% No italics, no superscripts
%% Use footnote or author note to identify equal contribution and/or contact author info

\begin{abstract}
  Marine videos present significant challenges for video understanding due to the dynamics of marine objects and the surrounding environment, camera motion, and the complexity of underwater scenes. Existing video captioning datasets, typically focused on generic or human-centric domains, often fail to generalize to the complexities of the marine environment and gain insights about marine life. To address these limitations, we propose a two-stage marine object-oriented video captioning pipeline. We introduce a comprehensive video understanding benchmark that leverages the triplets of video, text, and segmentation masks to facilitate visual grounding and captioning, leading to improved marine video understanding and analysis, and marine video generation. Additionally, we highlight the effectiveness of video splitting in order to detect salient object transitions in scene changes, which significantly enrich the semantics of captioning content. Our dataset and code have been released at \url{https://msc.hkustvgd.com}.
\end{abstract}




%   \includegraphics[width=\textwidth]{images/teaser2.pdf}
%   \Description{A teaser image showing MVK applications.}
%   % \caption{From top: Keyframes; Pixel-wise keyframe segmentation;conrvs Clip-level video summarization}
%   \caption{Overview of the \ourdataset dataset. The \ourdataset marine wildlife dataset is recorded across 13 countries, over 36 hours of marine video content. The dataset is associated with fine-grained annotations including clip-level textual descriptions provided by 18 biologists and pixel-level segmentation masks provided by 20 professionals. Our dataset is designed for marine video understanding and analysis, and marine video generation.}


\maketitle

\begin{figure}[ht]
    \centering
    \includegraphics[width=0.8\linewidth]{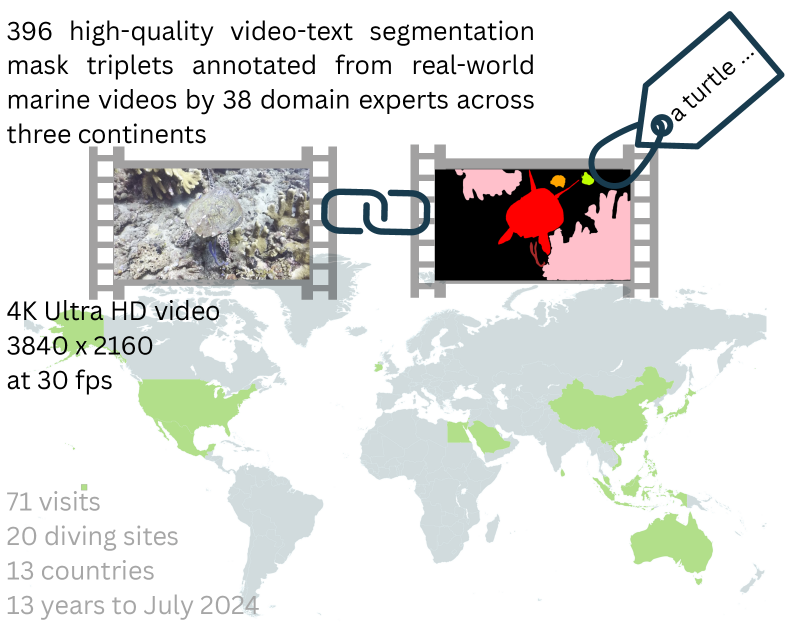}
    \caption{The \ourdataset dataset is recorded from 13 different countries, over 24.8 hours of marine video content. The dataset is associated with fine-grained annotations including clip-level textual descriptions provided by 18 biologists and pixel-level segmentation masks provided by 20 professionals.}
    \label{fig:worldmap}
    \vspace{-0.2in} % Uncomment if needed to adjust spacing
\end{figure}

\begin{figure*}
\centering
\includegraphics[width=1\linewidth]{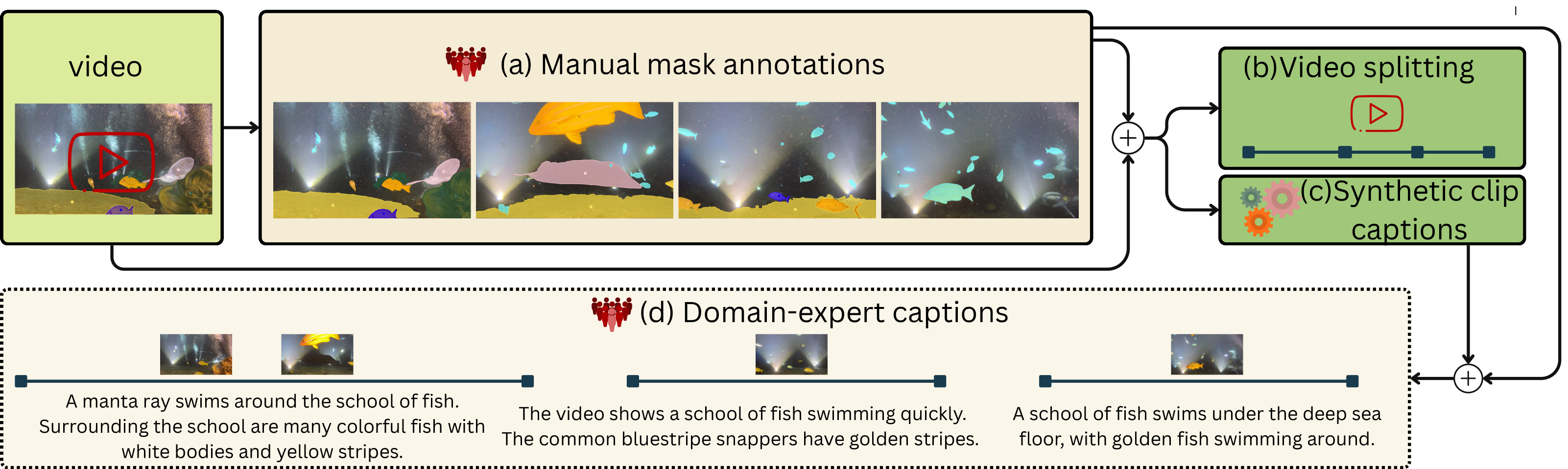}
   \caption{Annotation Generation. The dataset curation consists of two tasks: creating segmentation masks and captioning. (a) We manually label marine videos to identify the target objects. We aim to build a fine-grained marine wildlife video-text dataset, performing the following stages: (b) Extracting clips from marine videos using annotated images. (c) Using a VLM to generate synthetic clip-level captions. (d) Refining the synthetic captions by domain experts. Please zoom-in for the best view.}
   \label{fig:annotation-generation}
   \vspace{-0.1in}
\end{figure*}

\section{Introduction}
\label{sec:introduction}
The recent advent of Artificial Intelligence (AI) has led to a new research capacity in life and environmental sciences, from on-ground animal biology~\cite{stevens2024bioclip, brookes2025panaf, tuia2025mammalps} to marine biology, e.g., coral detection~\cite{zheng2024coralscop}, marine visual analysis~\cite{zheng2024marineinst, mehrab2024fish}. 

%However, there still remain significant challenges, e.g., labor-intensive data labeling of marine videos. 

The application of AI methods to ocean-related research is challenging due to the requirement of significant domain expertise and engineering work. Biologists have to collect data and manually label the data for specific marine species and tasks, then find and train a suitable model for each task. For real-world problems, off-the-shelf models often struggle to maintain comparative results. For example, visual grounding models, e.g., Grounding DINO~\cite{liu2023grounding} and SAM2~\cite{ravi2024sam}, are reliant on COCO's predefined classes~\cite{DBLP:conf/eccv/LinMBHPRDZ14}, and limited to user-defined input (e.g., customized text).
Due to the presence of hundreds of marine species, these approaches are not applicable to the marine domain. 
%To address this, open-set detection, for example, using Grounding DINO, relies on pre-trained CLIP~\cite{radford2021learning} model for concept generalization. This approach still struggles with detecting novel classes and fails to predict incorrect classes (e.g., a  classes videos or images). 

While Large Language Models (LLMs) can generate data at scale, they are prone to hallucination. Therefore, even with LLMs, human intervention is still needed to refine the synthetic data generated by the LLMs, especially in data-scarce domains. To address this issue, we propose a new marine video dataset, where the segmentation masks of marine objects are provided by annotators but the textual descriptions of the videos are generated automatically. Note that the MVK dataset introduced in~\cite{truong2023marine} also used LLMs to generate image-level captions. Compared with MVK, our \ourdataset provides more fine-grained and richer textual captions, generated from shorter temporal video segments (called clips) and validated by human experts. In addition, \ourdataset is two times larger than MVK. Video Browser Showdown utilized MVK for known item search in~\cite{vadicamo2024evaluating, rossetto2024results}, finding MVK a challenging benchmark dataset for visual KIS tasks.

%provides image-level captions generated by VLMs e.g., ClipCap~\cite{mokady2021clipcap} with raw videos. In contrast, \ourdataset is larger 2 times and we provide fine-grained, human-generated annotations, including instance video segmentation masks and clip-level captions. 

%\ourdataset consists of single-shot videos, each containing several clips or segments. Clips are defined based on the appearance of the target objects. 

%Video Browser Showdown uses MVK~\cite{truong2023marine} for known item search tasks. Specifically, a single video clip is randomly selected from the dataset, and participants need to find exact instances. There are three variants of Known Item Search (KIS), including visual KIS, textual KIS, and text-based KIS-C, which differ in their modal inputs (e.g., videos, text, or text with minimal initial details). ~\cite{vadicamo2024evaluating, rossetto2024results} indicate that MVK~\cite{truong2023marine} is a challenging benchmark dataset for visual KIS tasks. Our video dataset is recorded with a similar setting, and we are integrating MVK into our dataset to build a larger-scale video dataset.

Natural questions arise: \textit{Why do we create a new video captioning dataset? and why do we focus on clip-level captioning?} We argue that a clip-level video captioning dataset is essential for capturing richer semantic information, where each clip is defined as a semantically coherent segment. This leads to the development of our multimodal dataset comprising video-segmentation mask-text triplets, which explicitly link visual information with detailed textual descriptions and hold potential for multitask learning. Furthermore, we address the issue of hallucination prevalent in SOTA models, e.g., GPT-4.1~\cite{gpt}, suggesting that fine-grained, visually grounded data are crucial for generating accurate and reliable video captions.

In summary, our work makes the following contributions:
\begin{itemize}
\item We propose \ourdataset, the first real-world and large-scale marine video dataset, captured in various environments (see Figure~\ref{fig:worldmap}).
\item We provide high-quality video-segmentation mask-text triplets for short-time video clips using a two-stage video annotation pipeline (see Figure~\ref{fig:annotation-generation}).
\item We provide benchmark results on \ourdataset across applications, including video captioning, plot/clip-level captioning, video generation, and visual grounding.
\end{itemize}

\section{Related Work}
\label{sec:related_work}

\begin{figure*}
    \centering
    \includegraphics[width=0.8\linewidth]{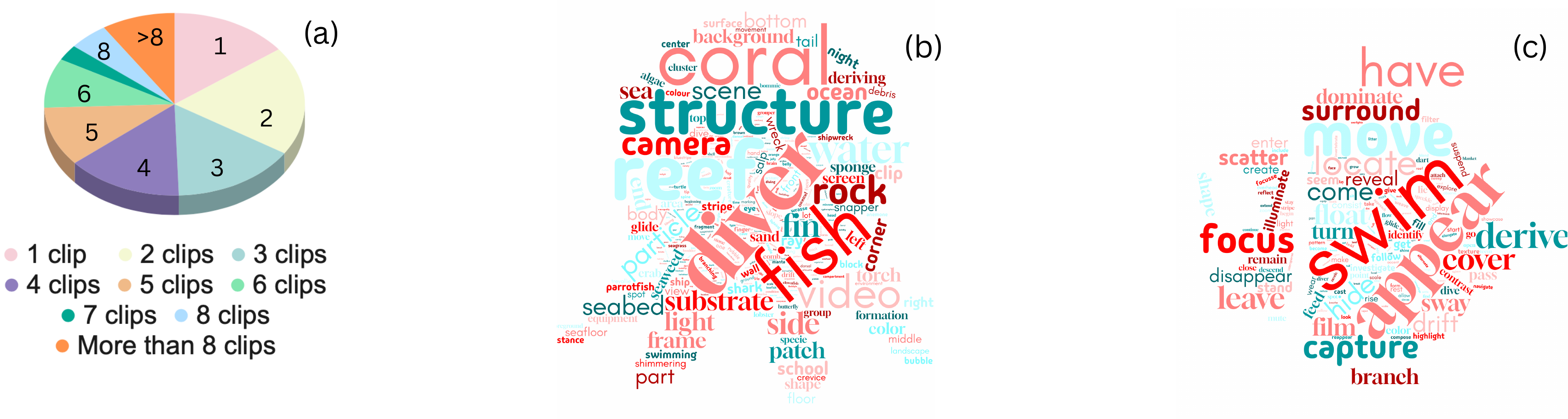}
    \caption{Overview of the \ourdataset dataset. (a) The dataset includes videos and clips, ranging from 1 to 30 clips per video. (b) and (c) Word cloud for noun and verb of marine captions generated by domain experts, respectively.}
    \label{fig:segment_wc_statistics}
    \vspace{-0.1in} % Uncomment if needed to adjust spacing
\end{figure*}

% \begin{figure*}
% \centering
% \includegraphics[width=0.9\linewidth]{images/mask_statistics.pdf}
%    \caption{\ourdataset Dataset. (a) The number of instances per category. (b) The average area of category. (c) Distribution of the number of instances by regions in the \ourdataset dataset.}
%    \label{fig:mask_statistics}
%    \vspace{-0.1in}
% \end{figure*}

\begin{figure*}
    \centering
    \begin{subfigure}[b]{0.3\textwidth}
        \scalebox{0.5}{%
            \begin{tikzpicture}
                \begin{axis}[
                    ybar,
                    bar width=20pt,
                    ylabel={Object count by category},
                    ylabel style={font=\Large},
                    xtick={1,2,3,4,5,6},
                    xticklabels={Sea floor, Reefs, Fish, Wrecks, Divers, Plants},
                    ymin=0,
                    enlarge x limits=0.15,
                    grid=both,
                    every grid/.style={line width=.1pt, draw=gray!50},
                    major grid style={line width=.2pt,draw=gray!80},
                    minor grid style={line width=.1pt,draw=gray!20},
                    ]
                    \addplot[fill=pink] coordinates {(1,6709) (2,2300) (3,2602) (4,232) (5,265) (6,151)};
                \end{axis}
            \end{tikzpicture}}
        \caption{}
        \label{fig:object_counts}
    \end{subfigure}%  <-- Important: Remove the space here
    \hfill%
    \begin{subfigure}[b]{0.3\textwidth}
        \scalebox{0.5}{%
            \begin{tikzpicture}
                \begin{axis}[
                    ybar,
                    bar width=20pt,
                    ylabel={Average Area by Category (\#pixels)},
                    xtick={1,2,3,4,5,6},
                    xticklabels={Sea floor, Reefs, Fish, Wrecks, Divers, Plants},
                    enlarge x limits=0.3,
                    ymin=0,
                    enlarge x limits=0.15,
                    grid=both,
                    every grid/.style={line width=.1pt, draw=gray!50},
                    major grid style={line width=.2pt, draw=gray!80},
                    minor grid style={line width=.1pt, draw=gray!20},
                    ytick={0, 200000, 400000, 600000, 800000, 1000000, 1200000}, % Set y-ticks
                    yticklabels={0, 200000, 400000, 600000, 800000, 1000000, 1200000} % Format y-tick labels
                    ]
                    \addplot[fill=teal] coordinates {(1,478387) (2,1081263) (3,55430) (4,820193) (5,92684) (6,760952)};
                \end{axis}
            \end{tikzpicture}}
        \caption{}
        \label{fig:average_areas}
    \end{subfigure}%  <-- Important: Remove the space here
    \hfill%
    \begin{subfigure}[b]{0.4\textwidth}
        \scalebox{0.55}{%
            \begin{tikzpicture}
                \begin{axis}[
                    xbar stacked,
                    width=\textwidth,
                    symbolic y coords={Tulamben, Dublin, TelAviv, BigIsland, Lajolle, WestPalm, Eqypt, HongKong},
                    ytick=data,
                    xmin=0,
                    xmax=1,
                    legend style={at={(0.5,-0.2)},anchor=north, font=\large, legend columns=6},
                    ]
                    \addplot[fill=blue!40] coordinates {(0.6747201596,HongKong) (0.2257551669,Eqypt) (0.1758241758,WestPalm) (0.2487046632,Lajolle) (0.1314285714,BigIsland) (0.2282157676,TelAviv) (0,Dublin) (0.2151898734,Tulamben)};
                    \addplot[fill=red!40] coordinates {(0.1282278621,HongKong) (0.559618442,Eqypt) (0.2454212454,WestPalm) (0.6632124352,Lajolle) (0.2857142857,BigIsland) (0.3692946058,TelAviv) (0.625,Dublin) (0.6329113924,Tulamben)};
                    \addplot[fill=green!40] coordinates {(0.1712290812,HongKong) (0.1717011129,Eqypt) (0.4013605442,WestPalm) (0.02590673575,Lajolle) (0.5542857143,BigIsland) (0.2821576763,TelAviv) (0,Dublin) (0.1518987342,Tulamben)};
                    \addplot[fill=orange!40] coordinates {(0.003657320182,HongKong) (0,Eqypt) (0.09576138148,WestPalm) (0,Lajolle) (0,BigIsland) (0.06639004149,TelAviv) (0,Dublin) (0,Tulamben)};
                    \addplot[fill=purple!40] coordinates {(0.01208023939,HongKong) (0.03497615262,Eqypt) (0.05965463108,WestPalm) (0.0103626943,Lajolle) (0.02857142857,BigIsland) (0.05394190871,TelAviv) (0,Dublin) (0,Tulamben)};
                    \addplot[fill=brown!40] coordinates {(0.01008533747,HongKong) (0.007949125596,Eqypt) (0.02197802198,WestPalm) (0.0518134715,Lajolle) (0,BigIsland) (0,TelAviv) (0.375,Dublin) (0,Tulamben)};

                    \legend{Sea floor, Reefs, Fish, Wrecks, Human divers, Aquatic plants}
                \end{axis}
            \end{tikzpicture}}
        \caption{}
    \end{subfigure}

    \caption{\ourdataset Dataset. (a) The number of instances per category. (b) The average area of category. (c) Distribution of the number of instances by regions in the \ourdataset dataset.}
    \label{fig:mask_statistics}
    \vspace{-0.15in}
\end{figure*}

%Several works exploited SAM~\cite{kirillov2023segment} for various marine applications. 

\paragraph{\textbf{Underwater Video Datasets.}}
Marine videos have recently attracted considerable attention from the computer vision community. MarineInst~\cite{ziqiang2024marineinst} introduced a large-scale marine dataset with instance segmentation masks, facilitating image-level visual analysis tasks. Only 10\% of the instance masks of MarineInst are annotated by humans while the remainder is generated by a segmentation model (e.g., SAM~\cite{kirillov2023segment}). CoralSCOP~\cite{zheng2024coralscop} demonstrates the effectiveness of SAM in coral image segmentation, specifically addressing over-segmentation issues. SAM has been commonly used to create marine image instance segmentation datasets such as Watermask~\cite{lian2023watermask} and USIS10K~\cite{lian2024diving}.

%\noindent\textbf{Video-Text Datasets.}
\paragraph{\textbf{Video-Text Datasets.}}
Recently, numerous large-scale video-text datasets, e.g., Koala-36M~\cite{wang2024koala}, HOIGen-1M~\cite{liu2025hoigen}, MiraData~\cite{ju2024miradata}, have leveraged automatic captioning systems (e.g., GPT-4V) to generate video captions. MovieBench~\cite{wu2024moviebench} provides video-, scene-, shot-level captions, also using GPT-4V, by incorporating additional semantic information. ViCaS~\cite{athar2024vicas} paves the way for new video-text datasets designed to evaluate video understanding tasks, including video captioning and visual grounding. Several datasets, e.g., BOVText~\cite{wu2021bilingual}, How2~\cite{sanabria2018how2}, and VALUE~\cite{li2021value} are utilized for tasks such as video retrieval, captioning, video text spotting, and video question answering. Furthermore, domain-specific datasets, e.g., BASKET~\cite{pan2025basket}, are useful in applications such as classification and video generation. Although existing video-text datasets play an important role for training downstream tasks, e.g., video generation, they rely on LLMs that often produce hallucinated outcomes human interventions.

%\noindent\textbf{Language-guilded segmentation datasets}
%In contrast, multi-object segmentation in datasets like MeViS further complicates the task, with each expression referring to an average of 1.59 objects. 

\paragraph{\textbf{Language-guided Segmentation Datasets.}} Early referring video object segmentation (RVOS) datasets, e.g., A2D Sentences~\cite{gavrilyuk2018actor}, J-HMDB Sentences~\cite{gavrilyuk2018actor}, DAVIS16-RVOS~\cite{khoreva2019video}, DAVIS17-RVOS~\cite{khoreva2019video}, and Refer-Youtube-VOS~\cite{seo2020urvos}, focus on single-object segmentation, where language prompts are used to describe a single target object. These datasets often feature expressions, describing static attributes like color and shape. The number of expressions per object varies across the datasets. For instance, A2D Sentences and J-HMDB Sentences typically have 1-2 expressions per object, while DAVIS17-RVOS boasts an average of 7.5 expressions per object, and MeViS~\cite{ding2023mevis} provides 3.5 expressions per object. However, it is observed that there is a lack of video grounding datasets. To our knowledge, \ourdataset is the first large-scale marine video text dataset for visual grounding. For visual grounding modeling, LLMs have been widely used to improve visual and language comprehension~\cite{rasheed2024glamm, he2024decoupling}. Grounding DINO~\cite{liu2023grounding}, built upon Transformers, is designed to generate multiple bounding boxes as prompts for SAM.

%LLM is employed to process the features extracted from dual vision encoders. Additionally, a spatio-temporal pixel decoder is used to predict output object masks that correspond to specific objects mentioned in the LLM output. 

\begin{figure*}[t]
    \centering
    \includegraphics[width=0.8\linewidth]{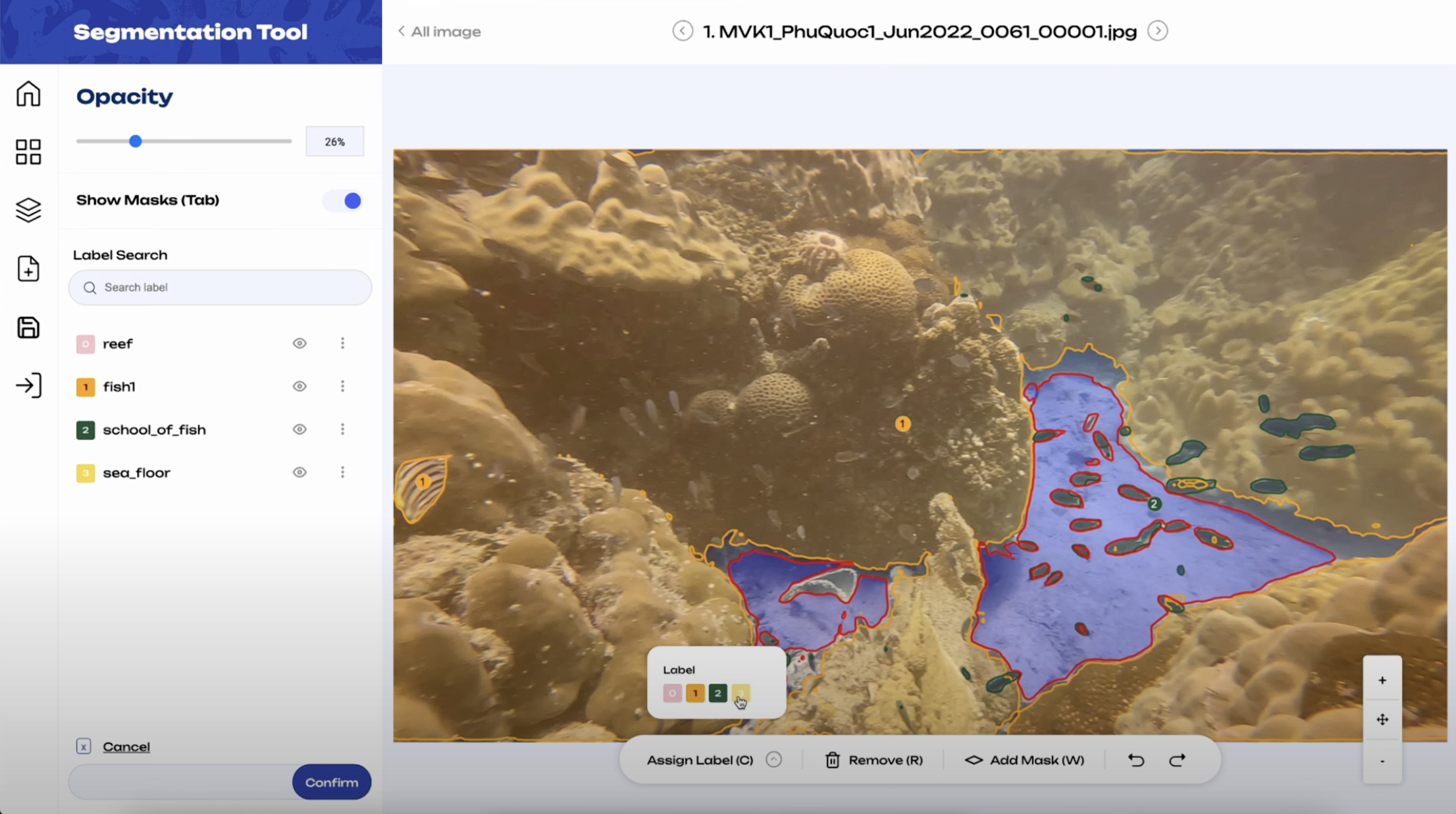}
    \caption{Visualization of our web-based segmentation annotation tool.}
    \label{fig:sat}
    \vspace{-0.1in}
\end{figure*}

\begin{figure}[t]
    \centering
    \includegraphics[width=0.85\linewidth]{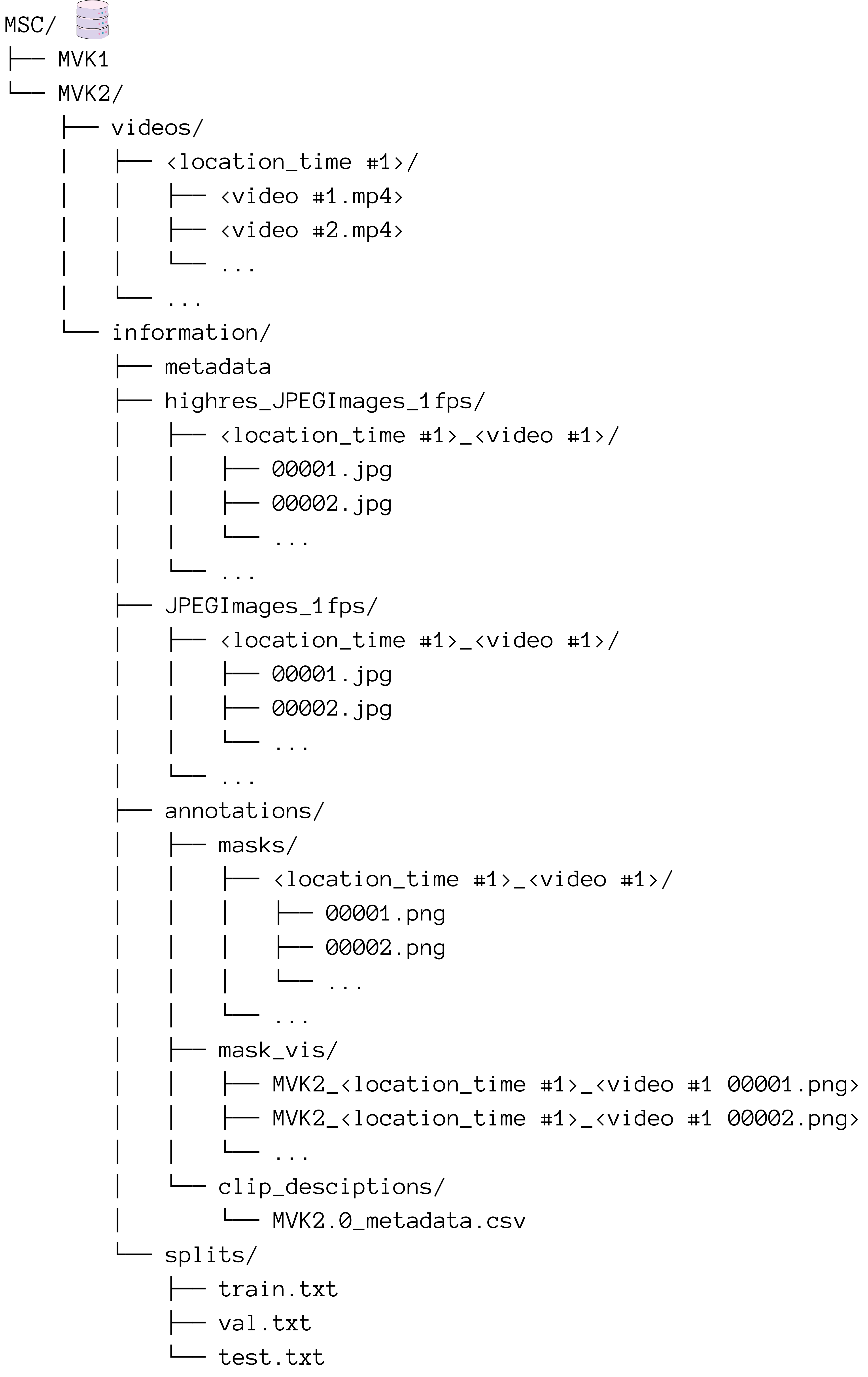}
    \caption{The \ourdataset's data organization.}
    \label{fig:data_structure}
    \vspace{-0.1in} 
\end{figure}

\section{Our \ourdataset}
\label{sec:dataset}

\subsection{Video Filtering}
\label{sec:video_filtering}

To choose marine videos for annotation, we considered three criteria (validated by human annotators): clarity, complexity, and diversity. (1) \textit{Clarity} means that only videos whose objects can be seen with moderate clarity are selected. (2) \textit{Complexity} means that videos containing objects in three or more distinct object types (e.g., fish, plants, reefs, divers) are selected. (3) \textit{Diversity} means that we choose videos with distinct scenes or objects among those of the same diving site. Applying the above criteria, we finally selected 396 videos out of 2,743 videos.

%Given the selected videos, we randomly sampled the top 396 videos out of 2743 videos that meet our criteria for diversity and visual clearness. Note that wildlife marine videos are recorded in diverse lighting condition and data labeling is often time-consuming and expensive. (1) Clearness filtering means that only objects can be seen with moderate clarity in selected videos. (2) Diversity filtering means that we select videos having diverse objects, having around three distinct object types (e.g., fish, plants, reefs, divers). (3) Dissimilarity filtering means that we have to choose distinct scene or objects among videos in the same diving site.

\subsection{Video Annotation}
\label{sec:annotation_process}

%\textbf{Video Collection.} To choose marine videos for annotation, we employ three filtering scores that require human annotators: clearness filtering, diversity filtering and dissimilarity filtering. We randomly sampled the top 396 videos out of 2743 videos that meet our criteria for diversity and visual clearness. Note that wildlife marine videos are recorded in diverse lighting condition and data labeling is often time-consuming and expensive. (1) Clearness filtering means that only objects can be seen with moderate clarity in selected videos. (2) Diversity filtering means that we select videos having diverse objects, having around three distinct object types (e.g., fish, plants, reefs, divers). (3) Dissimilarity filtering means that we have to choose distinct scene or objects among videos in the same diving site.

The annotation process involves two steps: creating video segmentation masks and captioning short-time clips. For the first step, annotators utilize a GUI tool to segment marine objects. We then use the manually annotated segmentation masks to identify target objects. In the second step, we provide high-quality clip-level captions using segmentation masks returned in the first step.

\paragraph{Step 1: Instance-level Video Segmentation.}
We developed a web-based annotation tool for marine video object segmentation in Fig. ~\ref{fig:sat}. This annotation tool inherits from SAM~\cite{kirillov2023segment} and \cite{wong2024coralscoplatlabelinganalyzingtool} to produce high-quality pixel-wise segmentation masks (that we call pseudo-masks), and then allows annotators to refine the generated masks in an iterative manner. The annotators finally provide a category for each segmented mask. We focus on six categories, including fish, reefs, aquatic plants, wrecks, human divers, and sea floor.

\paragraph{Step 2: Captioning.} We leveraged LLMs to generating captions for our collected videos. However, we observed that the generated captions of long-time videos are often superficial because of the lack of detailed descriptions for the events included in the videos. To address this issue, we split long-time videos into short-time clips, each of which captures a single-shot event. We found that this step helps enhance the semantics of the generated captions. 

%We also observed that user-provided text descriptions often capture the key events in marine videos, indicating that the task is inherently related to clip summarization and moment localization.
 
%The captioning consists of two stages: 1) extracting clips from marine videos (video splitting), and 2) generating captions for these clips. 

After splitting long-time videos into short-time clips, we used GPT-4.1~\cite{gpt}, Gemini-2.0 Flash-Lite~\cite{gemini}, Qwen-VL~\cite{Qwen-VL} to generate a textual description for each clip. The generated descriptions were then refined by biologists to reflect the semantic content specified by the segmented objects in the corresponding clips. The descriptions were refined to include the visual attributes and behavior (e.g., feeding, resting, breathing, social interactions, defense) of segmented objects and the background (surrounding environment) in the clips. Finally, the biologists produce a comprehensive caption for each video by aggregating the refined clip-level descriptions for the clips in that video, providing a concise summary of the video content.

% \subsection{Dataset Annotations}
% \label{sec:dataset_annotations}

% \textbf{Instance-level Video Segmentation.}

% \noindent\textbf{Video Captioning.}

% \subsection{Video Understanding with Grounded Segmentation and Clip-Level Captions.}
% \label{sec:video_understanding}

\begin{figure*}[t]
    \centering
    \includegraphics[width=1\linewidth]{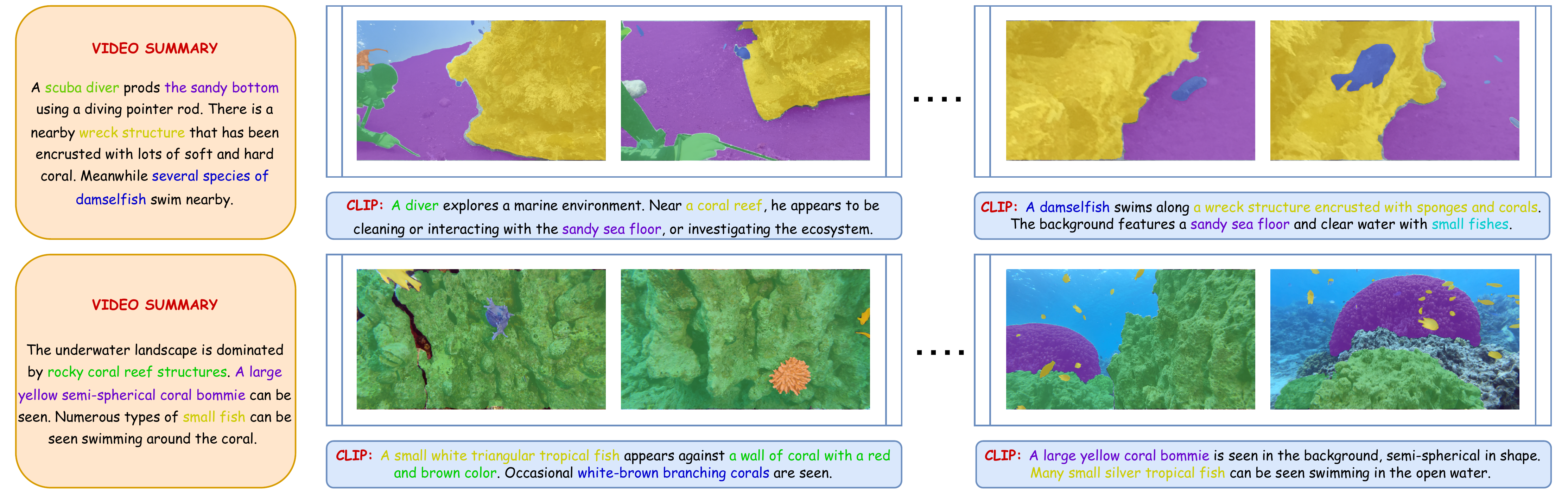}
    \caption{Visualization of video-text segmentation mask triplets.}
    \label{fig:examples}
    % \vspace{-0.1in} % Uncomment if needed to adjust spacing
\end{figure*}

\subsection{Analysis and Statistics}
\label{sec:dataset_analysis_statistics}
\noindent\textbf{Diverse Lighting Conditions and Scenarios.}
Our \ourdataset dataset was constructed by self-recording 71 visits across 20 distinct diving sites from 2011 to July 2024 (see Figure~\ref{fig:worldmap}). Additionally, \ourdataset is compiled with MVK \cite{truong2023marine}, which encompasses recordings from various regions and under diverse illumination conditions. Data acquisition for this dataset was performed using GoPro cameras. Unlike common video datasets gathered through web crawling, our focus was specifically on marine organisms captured during dedicated diving expeditions. Importantly, a consistent data acquisition setup was maintained throughout the entire collection process. We show our data’s directory structure in Figure~\ref{fig:data_structure}.

\begin{table}
    \centering
    \caption{Statistical overview of representative marine video datasets.}
    \scalebox{0.8}{%
    \begin{tabular}{l|p{2cm}|p{2cm}|c|c}
        \toprule
        Dataset   & Total Duration (hours) & Mean Duration (seconds) & \#Video & \#Country/Region \\ \midrule
        MVK \cite{truong2023marine} & 12.3          & 29.9         & 1378     & 11                \\
        Our \ourdataset & 24.8         & 32.8        & 2743     & 20                 \\ \bottomrule
    \end{tabular}
    \label{table:video_data_summary}}
    \vspace{-0.2in}
\end{table}

\begin{table*}
    \centering
    \caption{Video-level captioning on \ourdataset dataset.}
    \scalebox{1}{%
    \begin{tabular}{l|c|cccccc}
        \toprule
        Method      & Year      & BLEU$\uparrow$ & METEOR$\uparrow$ & ROUGE-L$\uparrow$ &  CIDEr$\uparrow$ & SPICE$\uparrow$ \\ \midrule		
        Qwen-VL-Chat 7B \cite{Qwen-VL}      & 2023  & 0.0000      & 0.0759      & 0.1384        & 0.0148     & 0.0848 \\
        LLaVA 7B \cite{llava2024}      & 2024  & \textbf{0.0125}      & 0.1241      & 0.1798        & 0.0000     & 0.0689 \\
        PLLaVA 7B \cite{xu2024pllava}      & 2024  & 0.0000	   & 0.1114      & 0.1443        & 0.0028     & 0.0533 \\
        Gemini-2.0 Flash-Lite~\cite{gemini}    &  2025              & 0.0000	& \textbf{0.1251}	& \textbf{0.1829}	& 0.0571	& 0.0870 \\
        % GPT-4.1      & 2025          & -      & -      & -        & -     & - \\
        % VideoRecap \cite{islam2024video}  & CVPR2024   & -      & -      & -        & -     & - \\
        \midrule
        MovieBench \cite{wu2024moviebench}    & 2025   & 0.0000	& 0.1213	& 0.1790	& \textbf{0.0898}	& \textbf{0.0914}  \\ \bottomrule
    \end{tabular}}
    \label{table:video_captioning}
    \vspace{-0.1in}
\end{table*}

\noindent\textbf{Diverse Descriptions.}
We observed that textual descriptions obtained from LLMs pre-trained on large-scale datasets often suffer from hallucinations. To address this issue, we integrate the segmentation masks of target objects into textual descriptions, enabling the descriptions to be more focused on the target objects' behaviors and the surrounding background, thereby eliminating hallucinations. This approach allows to create high-quality captions not only for visual grounding tasks but for the Text-to-Video (T2V) task via LLMs video captioning.

To further refine our synthetic captions, we engaged a team of 18 biologists to accurately identify marine organisms in the video footage. Synthetic captions generated by LLMs, i.e., GPT4.1~\cite{gpt}, QWen\cite{Qwen-VL}, Gemini 2.0 Flash-Lite~\cite{gemini}, are employed to reduce the workload for manual captioning. Each short-time clip in our dataset was annotated with 1 to 4 text descriptions, comprising 3 synthetic and 1 human-written caption. The distribution of annotated videos is illustrated in Figure ~\ref{fig:segment_wc_statistics} (a). The word cloud of text descriptions is illustrated in Figure ~\ref{fig:segment_wc_statistics} (b, c). 

\noindent\textbf{Challenges with Object Scale and Imbalance.}
As shown in Figure~\ref{fig:mask_statistics} (a), while the numbers of fish and coral reef over 2,000 each, fish are predominantly small objects (0.67\% of image area), whereas reefs are typically large 13.03\%. Additionally, human divers are less frequent but consistently small objects (1.12\% in Figure~\ref{fig:mask_statistics} (b)). Conversely, wrecks are rare but large objects (9.89\% in Figure~\ref{fig:mask_statistics} (a, b)). This highlights an imbalance in both object quantity and scale in \ourdataset dataset.

\section{Challenges}
\label{sec:challenges}

% \subsection{Interactive Search Tasks}
% TODO - describe VBS MVK interactive search task and cite VBS reports from 2024 and 2025... move it to introduction

%TODO: mention top systems and search approaches they used at VBS
\begin{table*}[ht]
    \centering
    \caption{Clip-level captioning on \ourdataset dataset.}
    \scalebox{1}{%
    \begin{tabular}{l|c|cccccc}
        \toprule
        Method             & Year                 & BLEU$\uparrow$ & METEOR$\uparrow$ & ROUGE-L$\uparrow$ &  CIDEr$\uparrow$ & SPICE$\uparrow$ \\ \midrule
        Qwen-VL-Chat 7B \cite{Qwen-VL}      & 2023                  &       0.0000	& 0.0832	& 0.1541	& 0.0046	& 0.0696   \\
        LLaVA 7B \cite{llava2024}              & 2023  &      0.0000	& 0.1337	& 0.1678	& 0.0057	& 0.0638        \\
        Gemini-2.0 Flash-Lite~\cite{gemini}    &  2025                  & 0.0931	& 0.1736	& 0.2890	& 0.3679	& 0.1647       \\
        GPT4.1~\cite{gpt}            &  2025              & \textbf{0.7196}	& \textbf{0.5186}	& \textbf{0.7844}	& \textbf{4.9314}	& \textbf{0.6139} \\
        % VideoRecap \cite{islam2024video}        & CVPR2024              &           &        &    & &       \\ 
        % \midrule
        % MovieBench \cite{wu2024moviebench}    & 2025   &  -         & -       & -     & - & -    \\
        \bottomrule
    \end{tabular}}
    \label{table:plot_level_captioning}
\end{table*}

\begin{table}
    \centering
    \caption{Visual Grounding performance on \ourdataset dataset.}
    \scalebox{0.9}{%
    \begin{tabular}{l|c|c|cc}
        \toprule
        Method                    & Year & Input & mIoU$\uparrow$   & Recall$\uparrow$  \\ \midrule
        GroundingDINO + SAM2     & 2024       & Labels                   & 0.6543 & 0.7135  \\
        GroundingDINO + SAM2     & 2024       & Caption                     & 0.4447 & 0.5244  \\
        GLaMM + SAM2              & 2024   & Caption                     &  0.6727    & 0.7255      \\
        VideoGLaMM        & 2025   & Caption                     & \textbf{0.6812}     & \textbf{0.7532}      \\
        \bottomrule
    \end{tabular}}
    \vspace{-0.2in}
\end{table}

\subsection{Video-level Captioning}
\label{sec:video_captioing}
Video captioning aims at generating a descriptive text for an input video sequence.

% Set up and discussion
\textbf{Baselines:} In this work, we evaluated several prevailing visual-language models (VLMs) for video-level captioning in the \ourdataset dataset. These models include Qwen-VL-Chat~\cite{Qwen-VL}, LLaVA~\cite{llava2024}, PLLaVA~\cite{xu2024pllava}, Gemini~\cite{gemini}, MovieBench~\cite{wu2024moviebench}. The models were used to generate captions at the video level for our marine video collection. 

% Model architecture
Qwen-VL uses Qwen as the language backbone and OpenCLIP ViT-bigG as the visual encoder, connected via a single-layer cross-attention module as the position-aware vision-language adapter. LLaVA employs a large language model, such as Vicuna, LLaMA, combined with a frozen vision encoder, such as CLIP ViT-L/14) via a single linear layer as a trainable projector. PLLaVA, an extension of LLaVA for videos, uses a parameter-free pooling strategy for video captioning. MovieBench is a data pipeline designed to create shot-level descriptions of scenes, with a particular emphasis on movie characters. 

% LLaVA-Next, Qwen2.5-VL: LLaVA-NeXT supports stronger LLM backbones (e.g., Qwen, LLaMA-3) and zero-shot video understanding.

% Set up
We set up to evaluate these VLMs with the guide prompt \textit{``Describe the video by following guidelines: you should give a paragraph with maximum 75 words; focus on the most obvious feature of the main objects in the initial frames; infer the behavior of the object (feeding, resting, breathing, social interactions, defense); and describe the background in about 10 words. Focus on fish, reefs, aquatic plants, wrecks, human divers, and sea floor. Omit the words `underwater' and `shows'.''}

%to guide the models to generate the caption as we predefine in problem formulation in Sec. \ref{sec:dataset}.

We utilized the lightweight versions (7B parameters) of the models to fit with our hardware specification NVIDIA RTX-3090 GPUs. Additionally, we amended MovieBench to generate detailed scene descriptions. Specifically, we input a list of keyframes from a video sequence, along with the images of target objects, and use GPT-4.1 to describe the observable features and behaviors of the target objects in no more than 75 words. The images of the target objects are originated from the segmentation masks in our dataset. To optimize the image token usage, each frame is resized to 540$\times$960. It is important to note that we omit the audio input required in the original MovieBench framework, as it is unavailable in our setup. We ultilised the captioning metrics, i.e., BLEU~\cite{papineni2002bleu}, METEOR~\cite{banerjee2005meteor}, ROUGE-L~\cite{lin2004rouge}, CIDEr~\cite{vedantam2015cider}, and SPICE~\cite{anderson2016spice}.

%video captioning pipelines used to generate large-scale video-text datasets, i.e., MovieBench~\cite{wu2024moviebench}, to make them work in the marine domain. Since the character bank used to detect individuals in a video frame is not available in our scenario, we replace the object detector in MovieBench~\cite{wu2024moviebench} with our object segmenter. Furthermore, as input dialogues and movie audio descriptions are unavailable for marine videos, these modalities are not exploited.

%We adopted the core methodology of MovieBench to generate detailed scene description. Specifically, we input a list of keyframes from a video sequence, along with the images of target objects, and use GPT-4.1 to describe the observable features and behaviors of the target objects in no more than 75 words. The images of the target objects are originated from the segmentation masks in our dataset. To optimize the image token usage, each frame is resized to 540$\times$960. It is important to note that we omit the audio input required in the original MovieBench framework, as it is unavailable in our setup.

\textbf{Analysis:}
Gemini-2.0 and MovieBench perform well across the benchmarking metrics. Specifically, Gemini achieves 0.1251 (METEOR) and 0.1829 (ROUGE), while MovieBench achieves 0.0898 (CIDEr) and 0.0914 (SPICE). MovieBench is based on GPT-4.1, we observe that commercial models often produce comparative results.

% Please refer to the supplementary for additional caption results. To evaluate the quality of generated captions, evaluating metrics BLEU \cite{papineni2002bleu}, METEOR \cite{banerjee2005meteor},	ROUGE-L \cite{lin2004rouge}, CIDEr \cite{vedantam2015cider}, and SPICE \cite{anderson2016spice} are utilized. These four metrics are usually used in machine translation task and captioning evaluation. We can see the results in table \ref{table:video_captioning}. We can easily find that these open source 7B pretrained model get the very low score. While the other commercial ones get better result. The results demonstrate that the model is capable of generating accurate captions that effectively describe scenes, including colors, observable traits, and behaviors. However, the model lacks domain-specific knowledge in marine biology, limiting its ability to provide more specialized descriptions. 

\subsection{Clip-level Captioning}
\label{sec:Plot_level_captioning}
Clip-level captioning aims to split a video into clips and then provide a caption for each clip. 

\textbf{Baselines:} We also leverage various open-source and commercial VLMs like in the above video level captioning on our \ourdataset dataset. With the similar setup to video captioning, we used the same prompt, but we input the frames from each clip rather than an entire video. We set the number of frames per clip to 10. To evaluate the clip-level captioning, we adopted the same metrics as in the Sec. \ref{sec:video_captioing}. 

% We follow the same MovieBench setup in Sec. \ref{sec:video_captioing}, but in the experiment, we input the video Clip, instead of the entire video.

% Models & setup
%\textbf{Baseline:} Qwen-VL-Chat \cite{Qwen-VL}, LLaVA \cite{llava2024}, Gemini-2.0 Flash-Lite \cite{team2023gemini}, GPT4.1 \cite{hurst2024gpt} are used to evaluate video captioning models at clip level for marine video collection. 

%Additionally, video captioning pipelines used to generate large-scale video-text datasets i.e., MovieBench \cite{wu2024moviebench} is reproduced on marine domain with some modifications. The Clip-level captioning settings of MovieBench\cite{wu2024moviebench} are the same with video captioning ones.

% Discussion
\textbf{Analysis:} 
Regarding clip-level captioning benchmark, GPT-4.1 demonstrates superior performance, achieving the highest scores across all benchmark metrics. Gemini-2.0 ranks as the second-best performer in this comparison. Specifically, GPT-4.1 achieved scores of 0.7196 (BLEU), 0.5186 (METEOR), 0.7844 (ROUGE-L), 4.9314 (CIDEr), and 0.6139 (SPICE), while Gemini 2.0 yielded respective scores of 0.0931 (BLEU), 0.1736 (METEOR), 0.2890 (ROUGE-L), 0.3679 (CIDEr), and 0.1647 (SPICE). Note, we observe commercial models consistently achieve competitive performance in video and clip level captioning on this challenging domain in Table ~\ref{table:plot_level_captioning}.

\subsection{Visual Grounding}
The visual grounding task in the \ourdataset dataset involves linking a target model’s response to a user-specific text query by identifying marine creatures, objects, and their behaviors. This task requires the model to understand both spatial and temporal aspects in visually complex marine scenes characterized by high object variability, occlusion, and challenging lighting.

Given a caption \textit{\{CAPTION\}} describing a clip, we prompt a target model using the format ``\textit{\{CAPTION\}. Please respond with segmentation masks}'' to extract segmentation maps aligned with the textual description. The expected output highlights relevant spatial regions across frames, demonstrating the model’s ability to understand both spatial details (what appears in each frame) and temporal dynamics (how things change over time). We evaluated the grounding quality using mIoU and Recall metrics across annotated frames.

\textbf{Baselines:} We first assessed the performance of an open-vocabulary model, GroundingDINO~\cite{liu2023grounding} combined with SAM2~\cite{ravi2024sam}. For further validation, we benchmarked recent LLM-based visual grounding models, including VideoGLaMM~\cite{munasinghe2024videoglamm} and GLaMM~\cite{rasheed2024glamm}. As GLaMM~\cite{rasheed2024glamm} is originally designed for image-based grounding, we extended it to the video domain by incorporating temporal capabilities through SAM2~\cite{ravi2024sam}.

\textbf{Analysis:} While GroundingDINO + SAM2 show promise, their reliance on COCO-style categories limits their generalization ability to the marine domain. Performance notably degrades with natural, unconstrained captions and only improves when explicit label names are extracted for prompting. In contrast, LLM-based models better support spatio-temporal reasoning, establishing a more robust foundation for grounding in complex marine environments.

\subsection{Text-to-video Generation}
\label{sec:t2v}
Our goal is to benchmark T2V models in generation of marine videos using clip-level captions from a video-text dataset. Pre-trained models were used in this experiment.

\textbf{Baselines:} We compared the open-source model Latte\cite{ma2025latte} and commercial models, i.e., Hailuo~\cite{hailuo}, Kling 1.5~\cite{klingai}, in marine video generation task. For the experimental setup, we used 50 clip-level caption prompts as input for the T2V models. \textbf{Analysis:}  We observe commercial models often perform well on video generation across T2V metrics in Table~\ref{tab:t2v}. Specifically, Hailuo gains the superior result with 0.3236 (CLIP-T) while Kling 1.5 gains 0.997 on temperal consistancy metric. Pre-trained T2V models exhibit suboptimal performance in FID and FVD metrics due to insufficient diversity of underwater visual content in existing training datasets.

\begin{table}
    \centering
    \caption{Evaluation results with open-source model and commercial models in video generation.}
    \scalebox{0.8}{
    \begin{tabular}{c|c|cccc}
        \toprule
        Method & Year & CLIP-T$\uparrow$ & Temp Consistency$\uparrow$ & FID$\downarrow$ & FVD$\downarrow$ \\ \midrule
        Latte\cite{ma2025latte} & 2025 & 0.3189 & 0.993 & 76.91 & 3123.05 \\
        \hline
        Hailuo~\cite{hailuo} & - & \textbf{0.3236} & 0.9934 & 83.68 & \textbf{2007.37} \\
        Kling 1.5~\cite{klingai} & 2024 & 0.3148 & \textbf{0.997} & \textbf{71.90} & 2820.24 \\
        \bottomrule
    \end{tabular}}
    \label{tab:t2v}
    \vspace{-0.2in}
\end{table}

\section{Conclusion}
\label{sec:conclusions}
This paper introduces \ourdataset, the first large-scale video dataset of marine wildlife. The dataset contains fine-grained annotations, including object segmentation masks, clip-level captions, and video summaries. We developed an effective two-stage data annotation pipeline minimizing hallucinations by LLMs. Our associated benchmark includes video captioning, visual grounding, and text-to-video generation. We believe that \ourdataset will contribute to facilitate research in marine video understanding.

\begin{acks}
We thank all human annotators who conducted the annotations and evaluations. We gratefully acknowledge Indo Ocean Foundation for providing invaluable biological insight. The work was partially supported by the Marine Conservation Enhancement Fund (MCEF20107 and MCEF22112), an internal grant from HKUST (R9429), HKUST Marine Robotics and Blue Economy Technology Grant. The authors would like to express their sincere gratitude to the “Sustainable Smart Campus as a Living Lab” (SSC) program at HKUST for its invaluable support. This program and staff not only provided essential funding and coordination, but also integrates sustainability into campus operations, which served as the living demonstration of the principles underpinning this research. The data of MSC is released for non-commercial research purposes only.
\end{acks}

\bibliographystyle{ACM-Reference-Format}
\bibliography{main}


\begin{thebibliography}{46}

%%% the Web address.

\ifx \showCODEN    \undefined \def \showCODEN     #1{\unskip}     \fi
\ifx \showISBNx    \undefined \def \showISBNx     #1{\unskip}     \fi
\ifx \showISBNxiii \undefined \def \showISBNxiii  #1{\unskip}     \fi
\ifx \showISSN     \undefined \def \showISSN      #1{\unskip}     \fi
\ifx \showLCCN     \undefined \def \showLCCN      #1{\unskip}     \fi
\ifx \shownote     \undefined \def \shownote      #1{#1}          \fi
\ifx \showarticletitle \undefined \def \showarticletitle #1{#1}   \fi
\ifx \showURL      \undefined \def \showURL       {\relax}        \fi
% invisible to TeX
\providecommand\bibfield[2]{#2}
\providecommand\bibinfo[2]{#2}
\providecommand\natexlab[1]{#1}
\providecommand\showeprint[2][]{arXiv:#2}

\bibitem[gem(mini)]%
        {gemini}
 \bibinfo{year}{Gemini}\natexlab{}.
\newblock \bibinfo{howpublished}{\url{https://gemini.google.com/}}.
\newblock


\bibitem[gpt(T 41)]%
        {gpt}
 \bibinfo{year}{GPT-4.1}\natexlab{}.
\newblock \bibinfo{howpublished}{\url{https://openai.com/index/gpt-4-1/}}.
\newblock


\bibitem[hai(iluo)]%
        {hailuo}
 \bibinfo{year}{Hailuo}\natexlab{}.
\newblock \bibinfo{howpublished}{\url{https://hailuoai.video}}.
\newblock


\bibitem[kli(g 15)]%
        {klingai}
 \bibinfo{year}{Kling 1.5}\natexlab{}.
\newblock \bibinfo{howpublished}{\url{https://klingai.com/}}.
\newblock


\bibitem[Anderson et~al\mbox{.}(2016)]%
        {anderson2016spice}
\bibfield{author}{\bibinfo{person}{Peter Anderson}, \bibinfo{person}{Basura Fernando}, \bibinfo{person}{Mark Johnson}, {and} \bibinfo{person}{Stephen Gould}.} \bibinfo{year}{2016}\natexlab{}.
\newblock \showarticletitle{Spice: Semantic propositional image caption evaluation}. In \bibinfo{booktitle}{\emph{Computer Vision--ECCV 2016: 14th European Conference, Amsterdam, The Netherlands, October 11-14, 2016, Proceedings, Part V 14}}. Springer, \bibinfo{pages}{382--398}.
\newblock


\bibitem[Athar et~al\mbox{.}(2024)]%
        {athar2024vicas}
\bibfield{author}{\bibinfo{person}{Ali Athar}, \bibinfo{person}{Xueqing Deng}, {and} \bibinfo{person}{Liang-Chieh Chen}.} \bibinfo{year}{2024}\natexlab{}.
\newblock \showarticletitle{ViCaS: A Dataset for Combining Holistic and Pixel-level Video Understanding using Captions with Grounded Segmentation}.
\newblock \bibinfo{journal}{\emph{arXiv preprint arXiv:2412.09754}} (\bibinfo{year}{2024}).
\newblock


\bibitem[Bai et~al\mbox{.}(2023)]%
        {Qwen-VL}
\bibfield{author}{\bibinfo{person}{Jinze Bai}, \bibinfo{person}{Shuai Bai}, \bibinfo{person}{Shusheng Yang}, \bibinfo{person}{Shijie Wang}, \bibinfo{person}{Sinan Tan}, \bibinfo{person}{Peng Wang}, \bibinfo{person}{Junyang Lin}, \bibinfo{person}{Chang Zhou}, {and} \bibinfo{person}{Jingren Zhou}.} \bibinfo{year}{2023}\natexlab{}.
\newblock \showarticletitle{Qwen-VL: A Versatile Vision-Language Model for Understanding, Localization, Text Reading, and Beyond}.
\newblock \bibinfo{journal}{\emph{arXiv preprint arXiv:2308.12966}} (\bibinfo{year}{2023}).
\newblock


\bibitem[Banerjee and Lavie(2005)]%
        {banerjee2005meteor}
\bibfield{author}{\bibinfo{person}{Satanjeev Banerjee} {and} \bibinfo{person}{Alon Lavie}.} \bibinfo{year}{2005}\natexlab{}.
\newblock \showarticletitle{METEOR: An automatic metric for MT evaluation with improved correlation with human judgments}. In \bibinfo{booktitle}{\emph{Proceedings of the acl workshop on intrinsic and extrinsic evaluation measures for machine translation and/or summarization}}. \bibinfo{pages}{65--72}.
\newblock


\bibitem[Brookes et~al\mbox{.}(2025)]%
        {brookes2025panaf}
\bibfield{author}{\bibinfo{person}{Otto Brookes}, \bibinfo{person}{Maksim Kukushkin}, \bibinfo{person}{Majid Mirmehdi}, \bibinfo{person}{Colleen Stephens}, \bibinfo{person}{Paula Dieguez}, \bibinfo{person}{Thurston~C Hicks}, \bibinfo{person}{Sorrel Jones}, \bibinfo{person}{Kevin Lee}, \bibinfo{person}{Maureen~S McCarthy}, \bibinfo{person}{Amelia Meier}, {et~al\mbox{.}}} \bibinfo{year}{2025}\natexlab{}.
\newblock \showarticletitle{The PanAf-FGBG Dataset: Understanding the Impact of Backgrounds in Wildlife Behaviour Recognition}.
\newblock \bibinfo{journal}{\emph{arXiv preprint arXiv:2502.21201}} (\bibinfo{year}{2025}).
\newblock


\bibitem[Ding et~al\mbox{.}(2023)]%
        {ding2023mevis}
\bibfield{author}{\bibinfo{person}{Henghui Ding}, \bibinfo{person}{Chang Liu}, \bibinfo{person}{Shuting He}, \bibinfo{person}{Xudong Jiang}, {and} \bibinfo{person}{Chen~Change Loy}.} \bibinfo{year}{2023}\natexlab{}.
\newblock \showarticletitle{MeViS: A large-scale benchmark for video segmentation with motion expressions}. In \bibinfo{booktitle}{\emph{Proceedings of the IEEE/CVF international conference on computer vision}}. \bibinfo{pages}{2694--2703}.
\newblock


\bibitem[Gavrilyuk et~al\mbox{.}(2018)]%
        {gavrilyuk2018actor}
\bibfield{author}{\bibinfo{person}{Kirill Gavrilyuk}, \bibinfo{person}{Amir Ghodrati}, \bibinfo{person}{Zhenyang Li}, {and} \bibinfo{person}{Cees~GM Snoek}.} \bibinfo{year}{2018}\natexlab{}.
\newblock \showarticletitle{Actor and action video segmentation from a sentence}. In \bibinfo{booktitle}{\emph{Proceedings of the IEEE conference on computer vision and pattern recognition}}. \bibinfo{pages}{5958--5966}.
\newblock


\bibitem[He and Ding(2024)]%
        {he2024decoupling}
\bibfield{author}{\bibinfo{person}{Shuting He} {and} \bibinfo{person}{Henghui Ding}.} \bibinfo{year}{2024}\natexlab{}.
\newblock \showarticletitle{Decoupling static and hierarchical motion perception for referring video segmentation}. In \bibinfo{booktitle}{\emph{Proceedings of the IEEE/CVF Conference on Computer Vision and Pattern Recognition}}. \bibinfo{pages}{13332--13341}.
\newblock


\bibitem[Ju et~al\mbox{.}(2024)]%
        {ju2024miradata}
\bibfield{author}{\bibinfo{person}{Xuan Ju}, \bibinfo{person}{Yiming Gao}, \bibinfo{person}{Zhaoyang Zhang}, \bibinfo{person}{Ziyang Yuan}, \bibinfo{person}{Xintao Wang}, \bibinfo{person}{Ailing Zeng}, \bibinfo{person}{Yu Xiong}, \bibinfo{person}{Qiang Xu}, {and} \bibinfo{person}{Ying Shan}.} \bibinfo{year}{2024}\natexlab{}.
\newblock \showarticletitle{Miradata: A large-scale video dataset with long durations and structured captions}.
\newblock \bibinfo{journal}{\emph{Advances in Neural Information Processing Systems}}  \bibinfo{volume}{37} (\bibinfo{year}{2024}), \bibinfo{pages}{48955--48970}.
\newblock


\bibitem[Khoreva et~al\mbox{.}(2019)]%
        {khoreva2019video}
\bibfield{author}{\bibinfo{person}{Anna Khoreva}, \bibinfo{person}{Anna Rohrbach}, {and} \bibinfo{person}{Bernt Schiele}.} \bibinfo{year}{2019}\natexlab{}.
\newblock \showarticletitle{Video object segmentation with language referring expressions}. In \bibinfo{booktitle}{\emph{Computer Vision--ACCV 2018: 14th Asian Conference on Computer Vision, Perth, Australia, December 2--6, 2018, Revised Selected Papers, Part IV 14}}. Springer, \bibinfo{pages}{123--141}.
\newblock


\bibitem[Kirillov et~al\mbox{.}(2023)]%
        {kirillov2023segment}
\bibfield{author}{\bibinfo{person}{Alexander Kirillov}, \bibinfo{person}{Eric Mintun}, \bibinfo{person}{Nikhila Ravi}, \bibinfo{person}{Hanzi Mao}, \bibinfo{person}{Chloe Rolland}, \bibinfo{person}{Laura Gustafson}, \bibinfo{person}{Tete Xiao}, \bibinfo{person}{Spencer Whitehead}, \bibinfo{person}{Alexander~C Berg}, \bibinfo{person}{Wan-Yen Lo}, {et~al\mbox{.}}} \bibinfo{year}{2023}\natexlab{}.
\newblock \showarticletitle{Segment anything}. In \bibinfo{booktitle}{\emph{Proceedings of the IEEE/CVF international conference on computer vision}}. \bibinfo{pages}{4015--4026}.
\newblock


\bibitem[Li et~al\mbox{.}(2021)]%
        {li2021value}
\bibfield{author}{\bibinfo{person}{Linjie Li}, \bibinfo{person}{Jie Lei}, \bibinfo{person}{Zhe Gan}, \bibinfo{person}{Licheng Yu}, \bibinfo{person}{Yen-Chun Chen}, \bibinfo{person}{Rohit Pillai}, \bibinfo{person}{Yu Cheng}, \bibinfo{person}{Luowei Zhou}, \bibinfo{person}{Xin~Eric Wang}, \bibinfo{person}{William~Yang Wang}, {et~al\mbox{.}}} \bibinfo{year}{2021}\natexlab{}.
\newblock \showarticletitle{Value: A multi-task benchmark for video-and-language understanding evaluation}.
\newblock \bibinfo{journal}{\emph{arXiv preprint arXiv:2106.04632}} (\bibinfo{year}{2021}).
\newblock


\bibitem[Lian et~al\mbox{.}(2023)]%
        {lian2023watermask}
\bibfield{author}{\bibinfo{person}{Shijie Lian}, \bibinfo{person}{Hua Li}, \bibinfo{person}{Runmin Cong}, \bibinfo{person}{Suqi Li}, \bibinfo{person}{Wei Zhang}, {and} \bibinfo{person}{Sam Kwong}.} \bibinfo{year}{2023}\natexlab{}.
\newblock \showarticletitle{Watermask: Instance segmentation for underwater imagery}. In \bibinfo{booktitle}{\emph{Proceedings of the IEEE/CVF International Conference on Computer Vision}}. \bibinfo{pages}{1305--1315}.
\newblock


\bibitem[Lian et~al\mbox{.}(2024)]%
        {lian2024diving}
\bibfield{author}{\bibinfo{person}{Shijie Lian}, \bibinfo{person}{Ziyi Zhang}, \bibinfo{person}{Hua Li}, \bibinfo{person}{Wenjie Li}, \bibinfo{person}{Laurence~Tianruo Yang}, \bibinfo{person}{Sam Kwong}, {and} \bibinfo{person}{Runmin Cong}.} \bibinfo{year}{2024}\natexlab{}.
\newblock \showarticletitle{Diving into underwater: Segment anything model guided underwater salient instance segmentation and a large-scale dataset}. In \bibinfo{booktitle}{\emph{Proceedings of the 41st International Conference on Machine Learning}}. \bibinfo{pages}{29545--29559}.
\newblock


\bibitem[Lin(2004)]%
        {lin2004rouge}
\bibfield{author}{\bibinfo{person}{Chin-Yew Lin}.} \bibinfo{year}{2004}\natexlab{}.
\newblock \showarticletitle{Rouge: A package for automatic evaluation of summaries}. In \bibinfo{booktitle}{\emph{Text summarization branches out}}. \bibinfo{pages}{74--81}.
\newblock


\bibitem[Lin et~al\mbox{.}(2014)]%
        {DBLP:conf/eccv/LinMBHPRDZ14}
\bibfield{author}{\bibinfo{person}{Tsung{-}Yi Lin}, \bibinfo{person}{Michael Maire}, \bibinfo{person}{Serge~J. Belongie}, \bibinfo{person}{James Hays}, \bibinfo{person}{Pietro Perona}, \bibinfo{person}{Deva Ramanan}, \bibinfo{person}{Piotr Doll{\'{a}}r}, {and} \bibinfo{person}{C.~Lawrence Zitnick}.} \bibinfo{year}{2014}\natexlab{}.
\newblock \showarticletitle{Microsoft {COCO:} Common Objects in Context}. In \bibinfo{booktitle}{\emph{Proceedings of the the European Conference on Computer Vision}}. \bibinfo{pages}{740--755}.
\newblock


\bibitem[Liu et~al\mbox{.}(2024a)]%
        {llava2024}
\bibfield{author}{\bibinfo{person}{Haotian Liu}, \bibinfo{person}{Chunyuan Li}, \bibinfo{person}{Yuheng Li}, {and} \bibinfo{person}{Yong~Jae Lee}.} \bibinfo{year}{2024}\natexlab{a}.
\newblock \showarticletitle{Improved baselines with visual instruction tuning}. In \bibinfo{booktitle}{\emph{Proceedings of the IEEE/CVF Conference on Computer Vision and Pattern Recognition}}. \bibinfo{pages}{26296--26306}.
\newblock


\bibitem[Liu et~al\mbox{.}(2025)]%
        {liu2025hoigen}
\bibfield{author}{\bibinfo{person}{Kun Liu}, \bibinfo{person}{Qi Liu}, \bibinfo{person}{Xinchen Liu}, \bibinfo{person}{Jie Li}, \bibinfo{person}{Yongdong Zhang}, \bibinfo{person}{Jiebo Luo}, \bibinfo{person}{Xiaodong He}, {and} \bibinfo{person}{Wu Liu}.} \bibinfo{year}{2025}\natexlab{}.
\newblock \showarticletitle{HOIGen-1M: A Large-scale Dataset for Human-Object Interaction Video Generation}.
\newblock \bibinfo{journal}{\emph{arXiv preprint arXiv:2503.23715}} (\bibinfo{year}{2025}).
\newblock


\bibitem[Liu et~al\mbox{.}(2024b)]%
        {liu2023grounding}
\bibfield{author}{\bibinfo{person}{Shilong Liu}, \bibinfo{person}{Zhaoyang Zeng}, \bibinfo{person}{Tianhe Ren}, \bibinfo{person}{Feng Li}, \bibinfo{person}{Hao Zhang}, \bibinfo{person}{Jie Yang}, \bibinfo{person}{Chunyuan Li}, \bibinfo{person}{Jianwei Yang}, \bibinfo{person}{Hang Su}, \bibinfo{person}{Jun Zhu}, {et~al\mbox{.}}} \bibinfo{year}{2024}\natexlab{b}.
\newblock \showarticletitle{Grounding dino: Marrying dino with grounded pre-training for open-set object detection}. In \bibinfo{booktitle}{\emph{arXiv preprint arXiv:2303.05499}}. \bibinfo{pages}{38--55}.
\newblock


\bibitem[Ma et~al\mbox{.}(2025)]%
        {ma2025latte}
\bibfield{author}{\bibinfo{person}{Xin Ma}, \bibinfo{person}{Yaohui Wang}, \bibinfo{person}{Xinyuan Chen}, \bibinfo{person}{Gengyun Jia}, \bibinfo{person}{Ziwei Liu}, \bibinfo{person}{Yuan-Fang Li}, \bibinfo{person}{Cunjian Chen}, {and} \bibinfo{person}{Yu Qiao}.} \bibinfo{year}{2025}\natexlab{}.
\newblock \showarticletitle{Latte: Latent diffusion transformer for video generation}.
\newblock \bibinfo{journal}{\emph{Transactions on Machine Learning Research}} (\bibinfo{year}{2025}).
\newblock


\bibitem[Mehrab et~al\mbox{.}(2024)]%
        {mehrab2024fish}
\bibfield{author}{\bibinfo{person}{Kazi~Sajeed Mehrab}, \bibinfo{person}{M Maruf}, \bibinfo{person}{Arka Daw}, \bibinfo{person}{Abhilash Neog}, \bibinfo{person}{Harish~Babu Manogaran}, \bibinfo{person}{Mridul Khurana}, \bibinfo{person}{Zhenyang Feng}, \bibinfo{person}{Bahadir Altintas}, \bibinfo{person}{Yasin Bakis}, \bibinfo{person}{Elizabeth~G Campolongo}, {et~al\mbox{.}}} \bibinfo{year}{2024}\natexlab{}.
\newblock \showarticletitle{Fish-vista: A multi-purpose dataset for understanding \& identification of traits from images}.
\newblock \bibinfo{journal}{\emph{arXiv preprint arXiv:2407.08027}} (\bibinfo{year}{2024}).
\newblock


\bibitem[Munasinghe et~al\mbox{.}(2024)]%
        {munasinghe2024videoglamm}
\bibfield{author}{\bibinfo{person}{Shehan Munasinghe}, \bibinfo{person}{Hanan Gani}, \bibinfo{person}{Wenqi Zhu}, \bibinfo{person}{Jiale Cao}, \bibinfo{person}{Eric Xing}, \bibinfo{person}{Fahad~Shahbaz Khan}, {and} \bibinfo{person}{Salman Khan}.} \bibinfo{year}{2024}\natexlab{}.
\newblock \showarticletitle{VideoGLaMM: A Large Multimodal Model for Pixel-Level Visual Grounding in Videos}.
\newblock \bibinfo{journal}{\emph{arXiv preprint arXiv:2411.04923}} (\bibinfo{year}{2024}).
\newblock


\bibitem[Pan et~al\mbox{.}(2025)]%
        {pan2025basket}
\bibfield{author}{\bibinfo{person}{Yulu Pan}, \bibinfo{person}{Ce Zhang}, {and} \bibinfo{person}{Gedas Bertasius}.} \bibinfo{year}{2025}\natexlab{}.
\newblock \showarticletitle{BASKET: A Large-Scale Video Dataset for Fine-Grained Skill Estimation}.
\newblock \bibinfo{journal}{\emph{arXiv preprint arXiv:2503.20781}} (\bibinfo{year}{2025}).
\newblock


\bibitem[Papineni et~al\mbox{.}(2002)]%
        {papineni2002bleu}
\bibfield{author}{\bibinfo{person}{Kishore Papineni}, \bibinfo{person}{Salim Roukos}, \bibinfo{person}{Todd Ward}, {and} \bibinfo{person}{Wei-Jing Zhu}.} \bibinfo{year}{2002}\natexlab{}.
\newblock \showarticletitle{Bleu: a method for automatic evaluation of machine translation}. In \bibinfo{booktitle}{\emph{Proceedings of the 40th annual meeting of the Association for Computational Linguistics}}. \bibinfo{pages}{311--318}.
\newblock


\bibitem[Rasheed et~al\mbox{.}(2024)]%
        {rasheed2024glamm}
\bibfield{author}{\bibinfo{person}{Hanoona Rasheed}, \bibinfo{person}{Muhammad Maaz}, \bibinfo{person}{Sahal Shaji}, \bibinfo{person}{Abdelrahman Shaker}, \bibinfo{person}{Salman Khan}, \bibinfo{person}{Hisham Cholakkal}, \bibinfo{person}{Rao~M Anwer}, \bibinfo{person}{Eric Xing}, \bibinfo{person}{Ming-Hsuan Yang}, {and} \bibinfo{person}{Fahad~S Khan}.} \bibinfo{year}{2024}\natexlab{}.
\newblock \showarticletitle{Glamm: Pixel grounding large multimodal model}. In \bibinfo{booktitle}{\emph{Proceedings of the IEEE/CVF Conference on Computer Vision and Pattern Recognition}}. \bibinfo{pages}{13009--13018}.
\newblock


\bibitem[Ravi et~al\mbox{.}(2024)]%
        {ravi2024sam}
\bibfield{author}{\bibinfo{person}{Nikhila Ravi}, \bibinfo{person}{Valentin Gabeur}, \bibinfo{person}{Yuan-Ting Hu}, \bibinfo{person}{Ronghang Hu}, \bibinfo{person}{Chaitanya Ryali}, \bibinfo{person}{Tengyu Ma}, \bibinfo{person}{Haitham Khedr}, \bibinfo{person}{Roman R{\"a}dle}, \bibinfo{person}{Chloe Rolland}, \bibinfo{person}{Laura Gustafson}, {et~al\mbox{.}}} \bibinfo{year}{2024}\natexlab{}.
\newblock \showarticletitle{Sam 2: Segment anything in images and videos}.
\newblock \bibinfo{journal}{\emph{arXiv preprint arXiv:2408.00714}} (\bibinfo{year}{2024}).
\newblock


\bibitem[Rossetto et~al\mbox{.}(2024)]%
        {rossetto2024results}
\bibfield{author}{\bibinfo{person}{Luca Rossetto}, \bibinfo{person}{Klaus Schoeffmann}, \bibinfo{person}{Cathal Gurrin}, \bibinfo{person}{Jakub Loko{\v{c}}}, {and} \bibinfo{person}{Werner Bailer}.} \bibinfo{year}{2024}\natexlab{}.
\newblock \showarticletitle{Results of the 2024 Video Browser Showdown}.
\newblock \bibinfo{journal}{\emph{arXiv preprint arXiv:2502.15683}} (\bibinfo{year}{2024}).
\newblock


\bibitem[Sanabria et~al\mbox{.}(2018)]%
        {sanabria2018how2}
\bibfield{author}{\bibinfo{person}{Ramon Sanabria}, \bibinfo{person}{Ozan Caglayan}, \bibinfo{person}{Shruti Palaskar}, \bibinfo{person}{Desmond Elliott}, \bibinfo{person}{Lo{\"\i}c Barrault}, \bibinfo{person}{Lucia Specia}, {and} \bibinfo{person}{Florian Metze}.} \bibinfo{year}{2018}\natexlab{}.
\newblock \showarticletitle{How2: a large-scale dataset for multimodal language understanding}.
\newblock \bibinfo{journal}{\emph{arXiv preprint arXiv:1811.00347}} (\bibinfo{year}{2018}).
\newblock


\bibitem[Seo et~al\mbox{.}(2020)]%
        {seo2020urvos}
\bibfield{author}{\bibinfo{person}{Seonguk Seo}, \bibinfo{person}{Joon-Young Lee}, {and} \bibinfo{person}{Bohyung Han}.} \bibinfo{year}{2020}\natexlab{}.
\newblock \showarticletitle{Urvos: Unified referring video object segmentation network with a large-scale benchmark}. In \bibinfo{booktitle}{\emph{Computer Vision--ECCV 2020: 16th European Conference, Glasgow, UK, August 23--28, 2020, Proceedings, Part XV 16}}. Springer, \bibinfo{pages}{208--223}.
\newblock


\bibitem[Stevens et~al\mbox{.}(2024)]%
        {stevens2024bioclip}
\bibfield{author}{\bibinfo{person}{Samuel Stevens}, \bibinfo{person}{Jiaman Wu}, \bibinfo{person}{Matthew~J Thompson}, \bibinfo{person}{Elizabeth~G Campolongo}, \bibinfo{person}{Chan~Hee Song}, \bibinfo{person}{David~Edward Carlyn}, \bibinfo{person}{Li Dong}, \bibinfo{person}{Wasila~M Dahdul}, \bibinfo{person}{Charles Stewart}, \bibinfo{person}{Tanya Berger-Wolf}, {et~al\mbox{.}}} \bibinfo{year}{2024}\natexlab{}.
\newblock \showarticletitle{Bioclip: A vision foundation model for the tree of life}. In \bibinfo{booktitle}{\emph{Proceedings of the IEEE/CVF conference on computer vision and pattern recognition}}. \bibinfo{pages}{19412--19424}.
\newblock


\bibitem[Truong et~al\mbox{.}(2023)]%
        {truong2023marine}
\bibfield{author}{\bibinfo{person}{Quang-Trung Truong}, \bibinfo{person}{Tuan-Anh Vu}, \bibinfo{person}{Tan-Sang Ha}, \bibinfo{person}{Jakub Loko{\v{c}}}, \bibinfo{person}{Yue-Him Wong}, \bibinfo{person}{Ajay Joneja}, {and} \bibinfo{person}{Sai-Kit Yeung}.} \bibinfo{year}{2023}\natexlab{}.
\newblock \showarticletitle{Marine Video Kit: A new marine video dataset for content-based analysis and retrieval}. In \bibinfo{booktitle}{\emph{International Conference on Multimedia Modeling}}. Springer, \bibinfo{pages}{539--550}.
\newblock


\bibitem[Tuia(2025)]%
        {tuia2025mammalps}
\bibfield{author}{\bibinfo{person}{Devis Tuia}.} \bibinfo{year}{2025}\natexlab{}.
\newblock \showarticletitle{MammAlps: A multi-view video behavior monitoring dataset of wild mammals in the Swiss Alps}. In \bibinfo{booktitle}{\emph{The IEEE/CVF Conference on Computer Vision and Pattern Recognition 2025}}.
\newblock


\bibitem[Vadicamo et~al\mbox{.}(2024)]%
        {vadicamo2024evaluating}
\bibfield{author}{\bibinfo{person}{Lucia Vadicamo}, \bibinfo{person}{Rahel Arnold}, \bibinfo{person}{Werner Bailer}, \bibinfo{person}{Fabio Carrara}, \bibinfo{person}{Cathal Gurrin}, \bibinfo{person}{Nico Hezel}, \bibinfo{person}{Xinghan Li}, \bibinfo{person}{Jakub Lokoc}, \bibinfo{person}{Sebastian Lubos}, \bibinfo{person}{Zhixin Ma}, {et~al\mbox{.}}} \bibinfo{year}{2024}\natexlab{}.
\newblock \showarticletitle{Evaluating performance and trends in interactive video retrieval: Insights from the 12th vbs competition}.
\newblock \bibinfo{journal}{\emph{IEEE Access}} (\bibinfo{year}{2024}).
\newblock


\bibitem[Vedantam et~al\mbox{.}(2015)]%
        {vedantam2015cider}
\bibfield{author}{\bibinfo{person}{Ramakrishna Vedantam}, \bibinfo{person}{C Lawrence~Zitnick}, {and} \bibinfo{person}{Devi Parikh}.} \bibinfo{year}{2015}\natexlab{}.
\newblock \showarticletitle{Cider: Consensus-based image description evaluation}. In \bibinfo{booktitle}{\emph{Proceedings of the IEEE conference on computer vision and pattern recognition}}. \bibinfo{pages}{4566--4575}.
\newblock


\bibitem[Wang et~al\mbox{.}(2024)]%
        {wang2024koala}
\bibfield{author}{\bibinfo{person}{Qiuheng Wang}, \bibinfo{person}{Yukai Shi}, \bibinfo{person}{Jiarong Ou}, \bibinfo{person}{Rui Chen}, \bibinfo{person}{Ke Lin}, \bibinfo{person}{Jiahao Wang}, \bibinfo{person}{Boyuan Jiang}, \bibinfo{person}{Haotian Yang}, \bibinfo{person}{Mingwu Zheng}, \bibinfo{person}{Xin Tao}, {et~al\mbox{.}}} \bibinfo{year}{2024}\natexlab{}.
\newblock \showarticletitle{Koala-36m: A large-scale video dataset improving consistency between fine-grained conditions and video content}.
\newblock \bibinfo{journal}{\emph{arXiv preprint arXiv:2410.08260}} (\bibinfo{year}{2024}).
\newblock


\bibitem[Wong et~al\mbox{.}(2024)]%
        {wong2024coralscoplatlabelinganalyzingtool}
\bibfield{author}{\bibinfo{person}{Yuk-Kwan Wong}, \bibinfo{person}{Ziqiang Zheng}, \bibinfo{person}{Mingzhe Zhang}, \bibinfo{person}{David Suggett}, {and} \bibinfo{person}{Sai-Kit Yeung}.} \bibinfo{year}{2024}\natexlab{}.
\newblock \bibinfo{title}{CoralSCOP-LAT: Labeling and Analyzing Tool for Coral Reef Images with Dense Mask}.
\newblock
\showeprint[arxiv]{2410.20436}~[cs.CV]
\urldef\tempurl%
\url{https://arxiv.org/abs/2410.20436}
\showURL{%
\tempurl}


\bibitem[Wu et~al\mbox{.}(2021)]%
        {wu2021bilingual}
\bibfield{author}{\bibinfo{person}{Weijia Wu}, \bibinfo{person}{Yuanqiang Cai}, \bibinfo{person}{Debing Zhang}, \bibinfo{person}{Sibo Wang}, \bibinfo{person}{Zhuang Li}, \bibinfo{person}{Jiahong Li}, \bibinfo{person}{Yejun Tang}, {and} \bibinfo{person}{Hong Zhou}.} \bibinfo{year}{2021}\natexlab{}.
\newblock \showarticletitle{A bilingual, openworld video text dataset and end-to-end video text spotter with transformer}.
\newblock \bibinfo{journal}{\emph{arXiv preprint arXiv:2112.04888}} (\bibinfo{year}{2021}).
\newblock


\bibitem[Wu et~al\mbox{.}(2024)]%
        {wu2024moviebench}
\bibfield{author}{\bibinfo{person}{Weijia Wu}, \bibinfo{person}{Mingyu Liu}, \bibinfo{person}{Zeyu Zhu}, \bibinfo{person}{Xi Xia}, \bibinfo{person}{Haoen Feng}, \bibinfo{person}{Wen Wang}, \bibinfo{person}{Kevin~Qinghong Lin}, \bibinfo{person}{Chunhua Shen}, {and} \bibinfo{person}{Mike~Zheng Shou}.} \bibinfo{year}{2024}\natexlab{}.
\newblock \showarticletitle{MovieBench: A Hierarchical Movie Level Dataset for Long Video Generation}.
\newblock \bibinfo{journal}{\emph{arXiv preprint arXiv:2411.15262}} (\bibinfo{year}{2024}).
\newblock


\bibitem[Xu et~al\mbox{.}(2024)]%
        {xu2024pllava}
\bibfield{author}{\bibinfo{person}{Lin Xu}, \bibinfo{person}{Yilin Zhao}, \bibinfo{person}{Daquan Zhou}, \bibinfo{person}{Zhijie Lin}, \bibinfo{person}{See~Kiong Ng}, {and} \bibinfo{person}{Jiashi Feng}.} \bibinfo{year}{2024}\natexlab{}.
\newblock \showarticletitle{Pllava: Parameter-free llava extension from images to videos for video dense captioning}.
\newblock \bibinfo{journal}{\emph{arXiv preprint arXiv:2404.16994}} (\bibinfo{year}{2024}).
\newblock


\bibitem[Zheng et~al\mbox{.}(2024a)]%
        {zheng2024marineinst}
\bibfield{author}{\bibinfo{person}{Ziqiang Zheng}, \bibinfo{person}{Yiwei Chen}, \bibinfo{person}{Huimin Zeng}, \bibinfo{person}{Tuan-Anh Vu}, \bibinfo{person}{Binh-Son Hua}, {and} \bibinfo{person}{Sai-Kit Yeung}.} \bibinfo{year}{2024}\natexlab{a}.
\newblock \showarticletitle{Marineinst: A foundation model for marine image analysis with instance visual description}. In \bibinfo{booktitle}{\emph{European Conference on Computer Vision}}. Springer, \bibinfo{pages}{239--257}.
\newblock


\bibitem[Zheng et~al\mbox{.}(2024b)]%
        {ziqiang2024marineinst}
\bibfield{author}{\bibinfo{person}{Ziqiang Zheng}, \bibinfo{person}{Yiwe Chen}, \bibinfo{person}{Huimin Zeng}, \bibinfo{person}{Tuan-Anh Vu}, \bibinfo{person}{Binh-Son Hua}, {and} \bibinfo{person}{Sai-Kit Yeung}.} \bibinfo{year}{2024}\natexlab{b}.
\newblock \showarticletitle{MarineInst: A Foundation Model for Marine Image Analysis with Instance Visual Description}. In \bibinfo{booktitle}{\emph{Proceedings of the European Conference on Computer Vision}}. \bibinfo{publisher}{Springer}.
\newblock


\bibitem[Zheng et~al\mbox{.}(2024c)]%
        {zheng2024coralscop}
\bibfield{author}{\bibinfo{person}{Ziqiang Zheng}, \bibinfo{person}{Haixin Liang}, \bibinfo{person}{Binh-Son Hua}, \bibinfo{person}{Yue~Him Wong}, \bibinfo{person}{Put Ang}, \bibinfo{person}{Apple Pui~Yi Chui}, {and} \bibinfo{person}{Sai-Kit Yeung}.} \bibinfo{year}{2024}\natexlab{c}.
\newblock \showarticletitle{CoralSCOP: segment any coral image on this planet}. In \bibinfo{booktitle}{\emph{Proceedings of the IEEE/CVF Conference on Computer Vision and Pattern Recognition}}. \bibinfo{pages}{28170--28180}.
\newblock


\end{thebibliography}

\end{document}